\documentclass{article}
\usepackage[latin9]{inputenc}
\usepackage{array}
\usepackage{rotating}
\usepackage{float}
\usepackage{multirow}
\usepackage{amsmath}
\usepackage{graphicx}
\usepackage{setspace}
\usepackage{microtype}
\usepackage{booktabs}
\usepackage{amssymb}
\usepackage{pifont}
\newcommand{\cmark}{\ding{51}}%
\newcommand{\xmark}{\ding{55}}%
\usepackage[unicode=true,
 bookmarks=false,
 breaklinks=false,pdfborder={0 0 1},backref=page,colorlinks=false]
 {hyperref}
\makeatletter

\usepackage{amsfonts}
\usepackage{subfig}

\usepackage[accepted]{icml2021}

\icmltitlerunning{Progressive Representative Labeling for Deep Semi-Supervised Learning}

\@ifundefined{showcaptionsetup}{}{%
 \PassOptionsToPackage{caption=false}{subfig}}
\usepackage{subfig}
\makeatother

\begin{document}

\twocolumn[
\icmltitle{Progressive Representative Labeling for Deep Semi-Supervised Learning}



\icmlsetsymbol{equal}{*}

\begin{icmlauthorlist}
\icmlauthor{Xiaopeng Yan}{equal,to}
\icmlauthor{Riquan Chen}{equal,to}
\icmlauthor{Litong Feng}{to}
\icmlauthor{Jingkang Yang}{to}
\icmlauthor{Huabin Zheng}{to}
\icmlauthor{Wayne Zhang}{to}
\end{icmlauthorlist}

\icmlaffiliation{to}{SenseTime Research, China}

\icmlcorrespondingauthor{Xiaopeng Yan}{sysuyanxp@gmail.com}

\icmlkeywords{Machine Learning, ICML}

\vskip 0.3in
]



\printAffiliationsAndNotice{\icmlEqualContribution} 

\begin{abstract}
Deep semi-supervised learning (SSL) has experienced significant attention in recent years, to leverage a huge amount of unlabeled data to improve the performance of deep learning with limited labeled data. Pseudo-labeling is a popular approach to expand the labeled dataset. However, whether there is a more effective way of labeling remains an open problem. In this paper, we propose to label only the most representative samples to expand the labeled set. Representative samples, selected by indegree of corresponding nodes on a directed k-nearest neighbor (kNN) graph, lie in the k-nearest neighborhood of many other samples. We design a graph neural network (GNN) labeler to label them in a progressive learning manner. Aided by the progressive GNN labeler, our deep SSL approach outperforms state-of-the-art methods on several popular SSL benchmarks including CIFAR-10, SVHN, and ILSVRC-2012. Notably, we achieve $72.1\%$ top-1 accuracy, surpassing the previous best result by $3.3\%$, on the challenging ImageNet benchmark with only $10\%$ labeled data.
\end{abstract}

\section{Introduction}
Deep neural networks~(DNNs) have been dominating the field of computer vision and even surpassed human-level performance for visual recognition~\cite{simonyan2014very,deng2009imagenet,he2016deep}. State-of-the-art visual recognition models for a wide range of tasks rely on supervised training, which requires large-scale human-labeled data. However, annotating data is expensive and sometimes involves expert knowledge. The expensive human annotation hinders the further development of data-hungry DNNs. Alternatively, semi-supervised learning (SSL)~\cite{zhu2005semi} leverages unlabeled data to improve a model's performance when only limited labeled data are available. As collecting large-scale unlabeled data is more practical and cheaper than labeled data, Deep SSL~\cite{sajjadi2016regularization,xie2019unsupervised,berthelot2019mixmatch,laine2016temporal,miyato2018virtual} has been an emerging research direction.

\begin{figure*}[tbh]
\begin{centering}
\subfloat[\label{fig:confidence}Sample selection by confidence]{\includegraphics[scale=0.38]{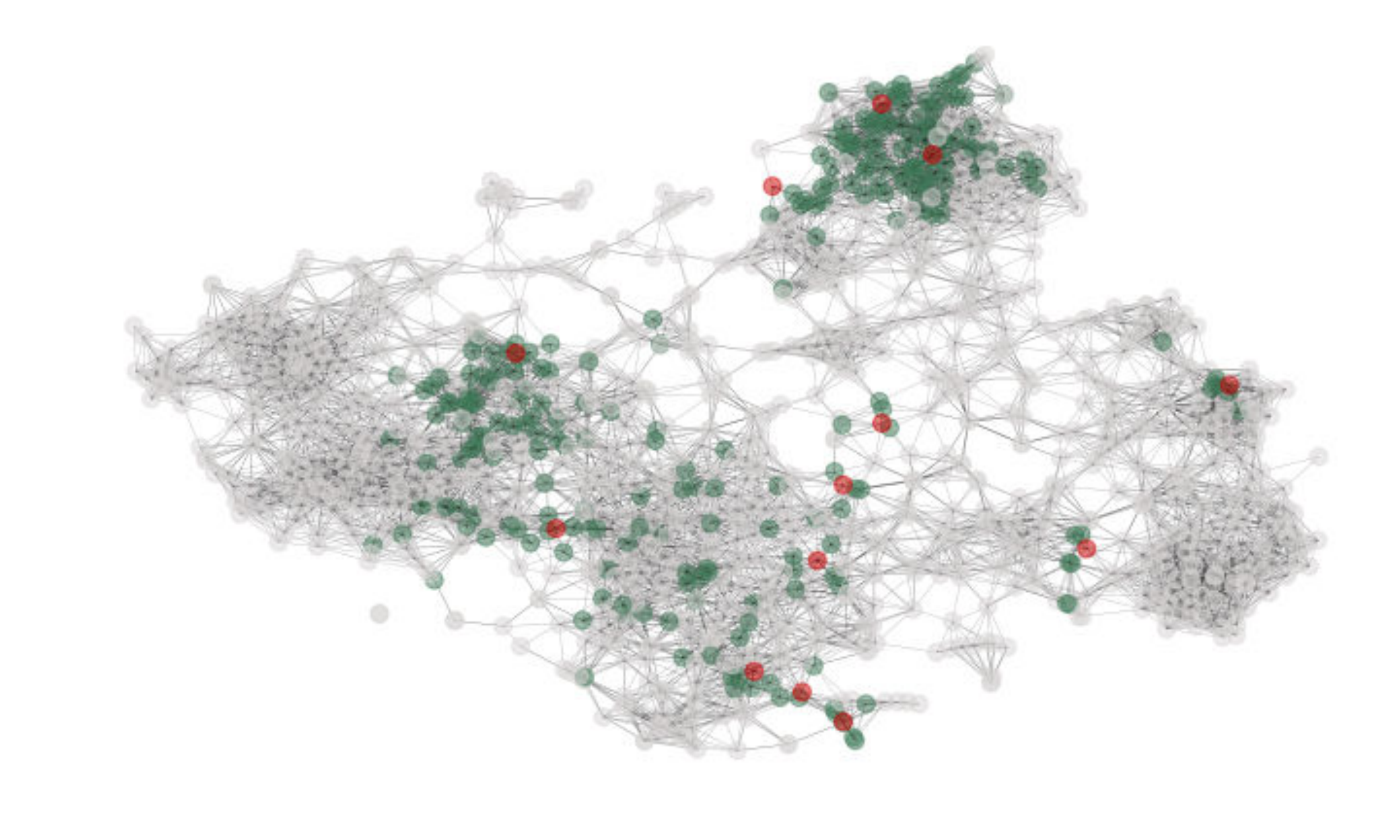}

}\subfloat[\label{fig:indegree}Sample selection by indegree]{\includegraphics[scale=0.38]{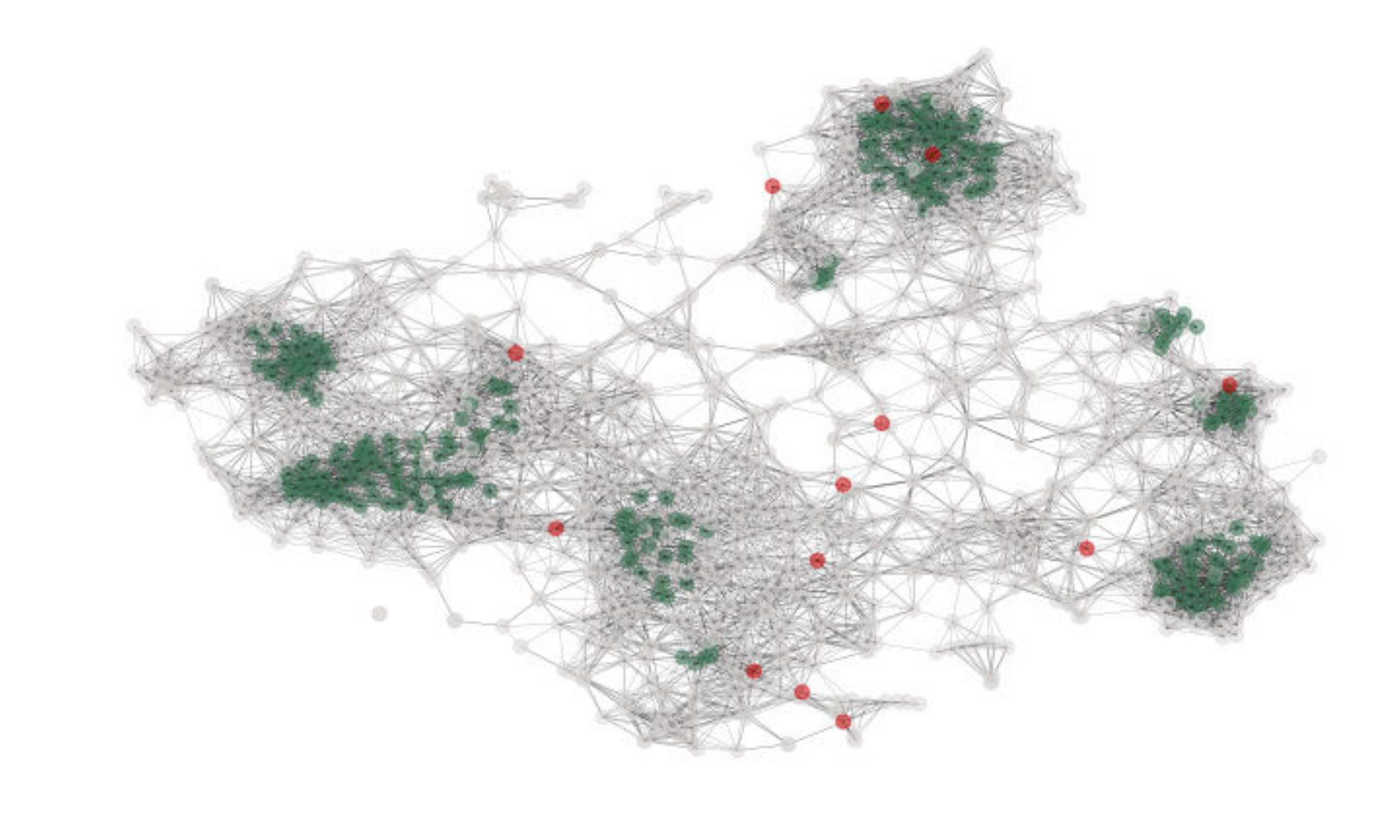}

}
\par\end{centering}
\caption{\label{fig:motivation}Samples selected by different samplers. Confidence sampler selects samples by ranking their confidence predicted by DNN, while indegree sampler sorts their indegrees on a constructed directed kNN graph. `Red' nodes represent labeled samples and `green' nodes represent selected samples. We can see that confidence sampler selects samples close to the labeled data, while indegree sampler selects representatives capturing the structure of the whole dataset.}
\vspace{1cm}
\end{figure*}

Pseudo-labeling is a simple but effective approach in Deep SSL. Previous approaches train an inductive (e.g., a DNN model~\cite{lee2013pseudo,yalniz2019billion,rosenberg2005semi}) or transductive (e.g., label propagation~\cite{iscen2019label,zhuang2019local}) model on the labeled set and pseudo-label the entire unlabeled set. We argue that these approaches have two limitations: 1) For sampling, most of these methods filter out unlabeled data by a hand-crafted confidence rule. As shown in Figure~\ref{fig:confidence}, unlabeled samples with high confidence are more likely to be close to the labeled data. Unfortunately, these samples are not representatives of the entire data distribution and such a sampling strategy is not the optimal way to capture the intrinsic structure of the whole unlabeled set. 2) For labeling, these approaches diffuse labels from the labeled data to the unlabeled data in a one-step manner. However, it seems extremely challenging to expand the labeling to the entire unlabeled data space, especially when the labeled samples are scarce. Thus, an effective sampling and labeling approach to utilize unlabeled data in deep SSL is awaiting exploration.

In this paper, we propose an indegree sampler to select the most representative samples for deep SSL as shown in Figure~\ref{fig:indegree}. 
As representative samples should be contiguous to as many samples as possible in the feature space, a directed k-nearest-neighbor (kNN) graph is constructed over all samples and then the samples (i.e., nodes in the graph) with high ranking by their indegrees are selected as representatives. With the indegree sampler, we can select the samples in the high-density region to capture the structure of the sample space. The \textit{boundary assumption}~\cite{chapelle2005semi} states that the decision boundary should not across the high-density region of a cluster. In other words, nodes lying in the high-density region are often reliable and are representative of the cluster. After sampling, we can apply an SSL approach to label the representatives.

For labeling, we employ graph neural networks~(GNNs)~\cite{kipf2016semi} for its popularity in graph-based SSL. GNNs nicely integrate feature extraction and graph topology in the design. To effectively propagate labels from the very few human-labeled samples, we train a GNN to label representatives and progressively expand the labeled training set. In detail, the following steps are performed repeatedly: First, representatives are selected by sorting the indegrees among all remaining unlabeled samples. Second, a GNN labeler is trained on the labeled set and makes predictions on the representatives. Finally, the representatives with high confidence are assigned by hard labels and added to the labeled set. For simplicity, we use two-layer simplified graph convolutional networks (SGC)~\cite{wu2019simplifying} as the GNN labeler. To save computational cost, we do not train the CNN feature extractor with the new labeled set, although the performance of deep SSL may be further improved.

To demonstrate the effectiveness of the proposed progressive representative labeling (PRL) approach, we apply state-of-the-art deep SSL methods~(e.g. consistency regularization) on the labeled set, pseudo-labeled representative set and remaining unlabeled set generated by PRL. Our deep SSL framework includes three stages: supervised DNN training, PRL\footnote{PRL is lightweight compared to the DNN training.}, and semi-supervised DNN finetuning.

Our main contributions are three folds:

1) We propose to pseudo-label representative samples for expanding the labeled set in deep SSL. Indegrees on a directed kNN graph are used for representatives selections.

2) We propose a GNN to label these representatives in a progressive learning manner. 

3) We demonstrate the effectiveness of the PRL approach with extensive experiments on several SSL benchmarks, including CIFAR-10, SHVN and ILSVRC-2012. Notably, we achieve $72.1\%$ top-1 accuracy, surpassing the previous best result by $3.3\%$, on the challenging ImageNet benchmark with only $10\%$ labeled data.

\section{Related Work}
Semi-supervised learning (SSL) is one of the classic problems in machine learning~\cite{zhu2005semi}. This section reviews the literature of deep SSL, an emerging topic in the deep learning era. The research of deep SSL can be divided into two major streams: \textit{pseudo-labeling}, and \textit{regularization}. 

\subsection{Pseudo-labeling for Deep SSL}
Pseudo-labeling methods~\cite{lee2013pseudo,yalniz2019billion,rosenberg2005semi} aim to take advantage of unlabeled data by assigning predicted labels to them. ~\cite{lee2013pseudo} infers pseudo-labels of unlabeled data by picking up the class with the largest probability and then fine-tunes the network with cross-entropy loss. Simply selecting the class with the largest probability is easy to bring noisy labels. To avoid this, ~\cite{yalniz2019billion} proposes a novel strategy of data sampling to help select reliable samples. The sampler first ranks the confidence within each individual class and then choose top-k samples for each class. ~\cite{xie2019self} infers noisy labels for unlabeled data and train a student model together with labeled data. Compared with these methods which do not take into account the importance of representative samples in SSL, our PRL approach pseudo-labels only the most representative samples among unlabeled data, selected with the largest indegree on the structure of data space modeled by a directed kNN graph. 

\subsection{Regularization for Deep SSL}
Regularization-based approaches, which optimize a heuristically-motivated objective, have been successful in deep SSL. Consistency regularization enforces that the model's output remains unchanged when the input is perturbed~\cite{sajjadi2016regularization, xie2019unsupervised, laine2016temporal}. Entropy Minimization~\cite{grandvalet2005semi, miyato2018virtual} encourages the model's output distribution to have low entropy (i.e., to make ``high-confidence'' predictions) on unlabeled data. MixMatch~\cite{berthelot2019mixmatch} also implicitly achieves entropy minimization through the use of a ``sharpening'' function on the target distribution for unlabeled data. Our PRL approach is complementary to regularization-based approaches, and we demonstrate its effectiveness with the simple consistency regularization~\cite{xie2019unsupervised}.

\subsection{Graph Neural Networks for Graph-based SSL} 
Graph neural networks (GNN) generalize convolutional neural networks to the graph domain~\cite{kipf2016semi, li2018deeper, wu2019simplifying, velivckovic2017graph}. The GNN model can naturally be applied to graph-based SSL, as it combines graph structures and node features in the convolution. In the GNN model, the features of unlabeled samples are mixed with those of nearby labeled samples, and propagated over the graph through multiple layers. In this paper, we choose the simplified graph convolutional networks~\cite{wu2019simplifying} for its computational efficiency. Note that other GNN can also be considered, such as graph attention networks~\cite{velivckovic2017graph}, but it is not the focus of this work.

\subsection{Label propagation for Deep SSL}
Recent approaches\cite{iscen2019label,yang2016revisiting,douze2018low,stretcu2019graph,thekumparampil2018attention,zhou2004learning} revisit the label propagation\cite{iscen2019label} algorithm to leverage unlabeled data. The goal of label propagation is to extend the labeled data via diffusing limited labeled data to the unlabeled data. \cite{douze2018low} employs label propagation on an approximate k-nearest neighbor  graph for few-shot learning. And \cite{liu2009robust} classifies the test images requiring the graph constructing on the entire dataset. \cite{luo2018smooth} proposes a graph as a similarity measure with respect to predicted features. Our work is different in that we inject a progressive scheme to train a GNN labeler by enlarging the labeled set using selected pseudo-labeled data with large indegree and high confidence, which are representative and reliable in the sample space. 

\section{Progressive Representative Labeling}
\label{method}
\begin{figure*}[t]
  \centering
  \includegraphics[width=1\linewidth]{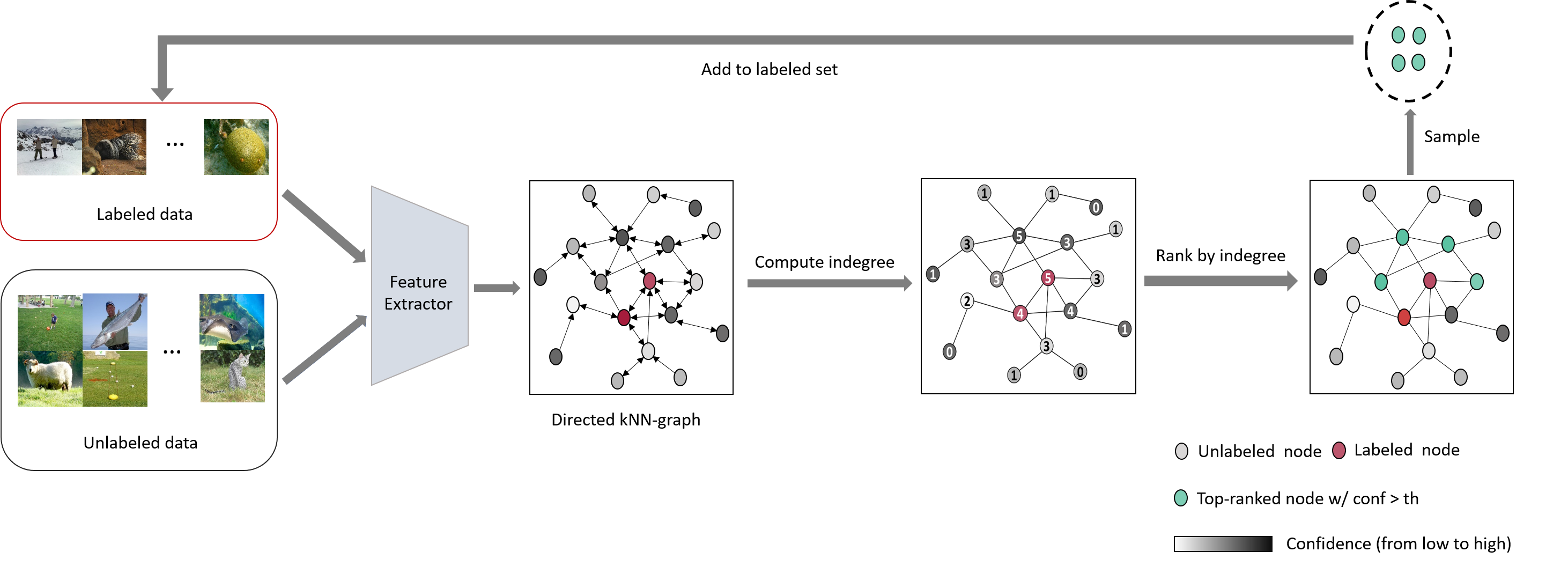}
  \caption{The overview of our proposed progressive representative labeling (PRL) based on indegree sampler. The grey line shows the progressive process of collecting representative unlabeled samples. Nodes in the directed kNN graph are sorted according to indegrees. Top-ranked nodes (samples) are added to $\mathcal{D}_{ru}$.}
  \label{fig:framework}
  \vspace{1cm}
\end{figure*}

In deep semi-supervised learning~(SSL), the learning algorithm has access to a small labeled set $\mathcal{D}_l = \{(x^l_1, y_1),(x^l_2, y_2),...,(x^l_n, y_n)\}$ along with a large unlabeled set $\mathcal{D}_u = \{x^u_1, x^u_2,..., x^u_m\}$ ($n \ll m$). The goal of our approach is to sample a representative unlabeled subset $\mathcal{D}_{ru}$ from $\mathcal{D}_u$, where the samples cover as much as possible area of entire data space (as shown in Figure\ref{fig:indegree}). To this end, we build a directed kNN graph with all data and characterize the density of the data by its indegree. Moreover, we propose a progressive representative labeling (PRL) approach to pseudo-label representative samples for expanding the labeled set as the Figure ~\ref{fig:framework}. We assign hard labels to samples in $\mathcal{D}_{ru}$ and combine advanced consistency training loss along with the cross-entropy loss on the $\mathcal{D}_l$, $\mathcal{D}_{ru}$, and the remaining non-representative unlabeled set $\mathcal{D}_{nu}$. 


\subsection{Representative Selection}

The \textit{boundary assumption}~\cite{chapelle2005semi} indicates that it is more likely that the decision boundary locates at the low-density region of a cluster, which means that nodes lying in the high-density region are often reliable. On the other hand, high-density nodes are also more representative of the cluster than those in the sparse region. 
Motivated by this observation, we propose a novel \textit{indegree sampler} to select representative unlabeled data. Concretely, a directed kNN graph is built based on the features of all samples in $\mathcal{D}_l \cup \mathcal{D}_u$, $\mathcal{G}=\{\mathbf{V}, \mathbf{A}, \mathbf{S}\}$, where $\mathbf{V}$ are nodes corresponding to samples, $\mathbf{A}$ is an adjacent matrix encoding graph edges, and $\mathbf{S}$ are nodes features. The direction of every edge is from a node to its top-k nearest neighbors, which can be single-directional or bi-directional. The edge construction of $\mathcal{G}$ can be expressed in the calculation of $\mathbf{A}$ as 
\begin{equation}
\mathbf{A}_{ij}=\left\{
\begin{aligned}
1, & v_j \in \mathcal{N}_k(v_i) \\
0, & v_j \notin \mathcal{N}_k(v_i) \\
\end{aligned}
\right.,
\end{equation}
where $\mathcal{N}_k(v_i)$ is the k-nearest neighbors of node $v_i$. The indegree of each node is the count of itself being neighbors of other nodes, which is presented as
\begin{equation}
    Indegree(v_i) = \sum_{j}\mathbf{A}_{ij}.
\end{equation}
All nodes are sorted according to indegrees. As shown in Figure~\ref{fig:indegree}, high indegree samples are among the kNN of other samples and lie in the high density region of the data space. Therefore, the indegree sampler helps selecting the most representative samples that can capture the intrinsic structure of the dataset. After selecting the nodes with large indegree, we employ a GNN labeler to pseudo-label those nodes. As shown in Figure~\ref{fig:motivation}, the selected samples by indegree sampler well cover the whole data space while the regular confidence thresholding sampler focuses merely on the samples around the labeled data..

\begin{algorithm}[!t]
  \caption{Progressive Representative Labeling}
  \label{alg:Representative Labeling}
  \begin{algorithmic}[1] 
    \STATE Build a kNN graph $\mathcal{G}$ with features extracted by a deep feature extractor such as CNN;
    \STATE Based on $\mathcal{G}$, train GNN labeler with $\mathcal{D}_l$;
    \STATE Assign running representative unlabeled set $\mathcal{D}_{ru} = \emptyset $;
    \WHILE{Size of $\mathcal{D}_{ru}$ $\textless$ Target}
    \STATE Calculate indegrees for nodes of $\mathcal{G}$;
    \STATE Sort nodes according to indegrees;
    \STATE Except samples in the running $\mathcal{D}_{ru}$, add selected samples into the running $\mathcal{D}_{ru} := \mathcal{D}_{ru}^{t}$;
    \STATE Finetune GNN labeler, update node features, update kNN graph $\mathcal{G}$;
    \ENDWHILE
    \STATE Pseudo-label $\mathcal{D}_{ru}$ with GNN labeler.
  \end{algorithmic}
\end{algorithm}
   \vspace{0.5cm}

\subsection{Representative Labeling with Progressive GNN}
As shown in Algorithm \ref{alg:Representative Labeling}, representative labeling is designed in a progressive learning manner. In first progressive step, the selected representative samples are supposed to be easy samples with reliable pseudo-labels. These reliable pseudo-labeled samples are supposed to be able to refine features and kNN graph $\mathcal{G}$, thus be helpful for mining hard representative unlabeled samples. 

\paragraph{GNN labeler} Graph neural network~(GNN)~\cite{scarselli2008graph} have been widely used in SSL for its superiority on modeling the graph-based relation. In this work, we employ GNN~\cite{scarselli2008graph} on graph $\mathcal{G}$ to predict pseudo-labels for representative samples. We adopt following steps to label representative samples. 1) First, we build the directed graph $\mathcal{G}$ using the features of $\mathcal{D}_l$. 2) Second, the GNN is trained the supervision of $\mathcal{D}_l$ with cross-entropy loss. 3) Finally, we assign hard labels for representative samples based on the GNN prediction and filter out the samples with confidence lower than threshold $\alpha$. 

Specifically, we employ a two-layer Simplified Graph Convolutional (SGC) network ~\cite{wu2019simplifying} as the GNN labeler. The SGC network takes the nodes feature and adjacent matrix as input and outputs new feature for each node. The forward computation in each layer of the SGC network is formulated as:
\begin{equation}
\mathbf{X}^* = (\hat{\mathbf{D}}^{-1/2}\hat{\mathbf{A}}\hat{\mathbf{D}}^{-1/2})\mathbf{X}\mathbf{\Theta}_g \\
\end{equation}
where $\hat{\mathbf{A}}=\mathbf{A} + \mathbf{I}$, and $\hat{\mathbf{D}}$ is a diagonal matrix with diagonal entries as $\hat{\mathbf{D}}_{ii}=\sum_{j}\hat{\mathbf{A}}_{ij}$. $\mathbf{X} \in \mathbb{R}^{n\times n}, \mathbf{X}^* \in \mathbb{R}^{n\times d^\prime}$ are the input and output of SGC layer, $\Theta_g \in \mathbb{R}^{d \times d^\prime}$ are learnable parameters. 

\paragraph{Progressive GNN} The above shows the first labeling process by GNN. We then plug the GNN into a progressive process. Concretely, we denote the GNN labeler in iteration $t$ as $G^t$ and the input feature $\mathbf{f}^t_i$ for node $i$ is computed as:
\begin{equation}
\mathbf{f}^t_i=\left\{
\begin{aligned}
& \phi(x_i), & t = 0\\
& G^{t-1}(\mathbf{f}^{t-1}_i), & t > 0\\
\end{aligned}
\right.
\end{equation}
where $\phi$ is a CNN model trained on $\mathcal{D}_l$, $x_i$ is the image of node $i$. Then we train the GNN by the expanded set $\mathcal{D}_l \cup \mathcal{D}_{ru}^{0} \cup, \dots,\cup, \mathcal{D}_{ru}^{t-1}$, where $\mathcal{D}_{ru}^{0}, ..., \mathcal{D}_{ru}^{t-1}$ is the representative set collected from iteration 0 to iteration $t-1$. A new set $\mathcal{D}_{ru}^t$ will be selected from the remaining unlabeled set. We repeat the procedure for several time until the size of $\mathcal{D}_{ru} = \mathcal{D}_{ru}^{0} \cup, \dots,\cup, \mathcal{D}_{ru}^{T}$ ($T$ is the total number of progressive iterations) reaches its target. Instead of labeling all samples simultaneously, the GNN gradually enlarge the labeled set by adding nodes from high-density (certain) region to low-density (certain) region in a progressive learning manner. Meanwhile, it is more accurate to infer the low-density nodes when gradually adding high-density nodes to train the GNN.

\subsection{Our Deep SSL Framework}
The pipeline of our framework is shown in Algorithm ~\ref{alg:SSL approach}. First, we train a DNN model as a feature extractor on $\mathcal{D}_l$. Second, our representative labeling approach utilizes a progressive GNN to collect and pseudo-label a representative set $\mathcal{D}_{ru}$. Finally, we finetune the model by consistency loss along with the cross-entropy loss by using $\mathcal{D}_l$, $\mathcal{D}_{ru}$ and the remaining unlabeled set $\mathcal{D}_{nu}$. The objective function in this phase is formulated as:

\begin{equation}
\begin{aligned}
loss  = &\sum_{(x, y)\in \mathcal{D}_l\cup \mathcal{D}_{ru}} {\rm{CE}} (x, y, \mathbf{p}_x) \\
& + \sum_{x\in \mathcal{D}_{nu}} {\rm{KLD}}(\mathbf{p}_x, \hat{\mathbf{p}}_x) \\
\label{eq:loss_Pseudo-labeling+}
\end{aligned}
\end{equation}
where $\mathbf{p}_x$ and $\hat{\mathbf{p}}_x$ are the probability vector of sample $x$ with a weak augmentation (e.g., random crop) and strong augmentation (e.g., random augment). ${\rm{CE}}$ and ${\rm{KLD}}$ means the cross entropy loss and Kullback-Leibler divergence loss function, defined as following:
\begin{equation}
\begin{aligned}
{\rm{CE}} (x, y, \mathbf{p}_x) & = -\sum_{k}\mathbf{1}(k=y)\log(\mathbf{p}_x^k) \\
{\rm{KLD}}(\mathbf{p}_x, \hat{\mathbf{p}}_x) & = -\sum_{k} \mathbf{p}_x^k \log \frac{\hat{\mathbf{p}}_x^k}{\mathbf{p}_x^k}
\label{eq:loss_Pseudo-labeling_each}
\end{aligned}
\end{equation}
where $\mathbf{p}_x^k$ and $\hat{\mathbf{p}}_x^k$ are the probalitity of the $k$th category. In fact, the PRL is a lightweight component compared to the initial DNN training and final model finetuning and is flexible to any semi-supervised framework.

\begin{algorithm}[h]
  \caption{Deep SSL framework}
  \label{alg:SSL approach}
  \begin{algorithmic}[1] 
    \STATE Train an initial CNN with labeled data $\mathcal{D}_l$, as the feature extractor;
    \STATE Build an directed kNN graph $\mathcal{G}$ with extracted features on labeled data $\mathcal{D}_l$ and unlabeled data $\mathcal{D}_u$;
    \STATE Collect representative unlabeled set $\mathcal{D}_{ru}$ using the proposed indegree-sampler and progressively train a GNN labeler;
    \STATE Pseudo-label $\mathcal{D}_{ru}$ with the GNN labeler;
    \STATE Finetune the CNN model with $\mathcal{D}_l$, $\mathcal{D}_{ru}$ and $\mathcal{D}_{nu}$ as Equation (\ref{eq:loss_Pseudo-labeling+})
    \end{algorithmic}
\end{algorithm}
\vspace{0.5cm}

\section{Experiments}

\subsection{Datasets} 
In the semi-supervised learning setting, the entire training set will be split into two parts. 
A small portion of training images are treated as labeled data and the rest are as unlabeled data. 
We conduct experiments on the following three standard semi-supervised image classification benchmarks:
\begin{itemize}
    \item For CIFAR-10~\cite{krizhevsky2010convolutional}, we provide both settings of 1\% labeled set and 10\% labeled set. Under 1\% labeled setting, we have 250 labeled data and 49,750 unlabeled data, with each class having only 25 labeled data. For 10\% labeled setting, the number of labeled and unlabeled data reaches 4,000 and 46,000, respectively.
    
    \item Street View House Numbers (SVHN)~\cite{netzer2011reading} gets a similar settings with CIFAR-10, as we have 250 labeled data and 72,007 unlabeled data in the setting of SVHN~(250), and 1,000 labeled with 71,257 unlabeled data for the setting of SVHN~(1000).
    
    \item ILSVRC-2012~\cite{deng2009imagenet} is adapted for semi-supervised learning setting following previous work~\cite{zhai2019s4l,yalniz2019billion}. We use 10\% of the labeled data~(roughly 12,8000 samples) of the ImageNet~\cite{deng2009imagenet} dataset and use the rest samples as unlabeled data.
\end{itemize}

\subsection{Implementation Details} 
Following previous works~\cite{xie2019unsupervised,berthelot2019mixmatch,tarvainen2017mean,miyato2018virtual}, we employ Wide-ResNet-28-2~\cite{zagoruyko2016wide} as the base model for CIFAR-10~\cite{krizhevsky2010convolutional} and SVHN~\cite{netzer2011reading}, and ResNet-50~\cite{he2016deep} for ImageNet~\cite{deng2009imagenet}. In the progressive representative labeling~(PRL), we consider k=5 neighbors for each sample to construct the kNN graph and we stack two SGC layers as the SGC labeler where the dimension of the hidden state is set as 64. Besides, we progressively train the PRL for $T=3$ iterations and the indegree sampler selects 30\%, 40\%, and 50\% representative samples by indegree in the order of most to least with a confidence threshold $\alpha=0.5$ for each iteration step, respectively. In the next section, we will analyze the progressive iterations $T$ in details. For fair comparison, we use the same training protocols, including data preprocessing, learning rate schedule and the optimizer across all SSL methods. In details, we implement our progressive representative labeling approach in PyTorch~\cite{paszke2017automatic} running with TITAN X GPU. With one GPU, our PRL costs only half an hour with Faiss~\cite{johnson2019billion} tools for billion-scale nearest neighbour search, which is negligible compared to finetuning the CNN model with 8 GPUs that costs nearly 24 hours. 

\begin{table*}[htb]
      \center
      \setlength{\tabcolsep}{2pt}
         \caption{Top-1 and top-5 accuracy on ImageNet~\cite{deng2009imagenet} validation set with only 10\% labeled data}
         \label{tab:imagenet}
         \vspace{6pt}
      \fontsize{9}{10}{\selectfont
         \begin{tabular}{@{\hskip 0.1in}l@{\hskip 0.1in}c@{\hskip 0.1in}c@{\hskip 0.1in}c@{\hskip 0.1in}c@{\hskip 0.1in}c@{\hskip 0.1in}c@{\hskip 0.1in}c@{\hskip 0.1in} c@{\hskip 0.1in} c@{\hskip 0.1in}}
            \toprule
             & Supervised & Pseudo-labels & VAT  & VAT-EM & $S^4L$ & LLP & UDA~(w/ Aug) & Ours \\
            \midrule
            Top-1 Accuracy &- &- &-  &- &- &- &68.78 &\textbf{72.08}\\
            Top-5 Accuracy &80.43 &82.41 &82.78 &83.39 &83.83 &88.53 &88.80 &\textbf{90.75}\\
            \bottomrule
         \end{tabular}}
      \end{table*}
         \vspace{1cm}

\begin{table*}[h]
      \center
      \setlength{\tabcolsep}{2.5pt}
      \fontsize{9}{10}{\selectfont
        \caption{Test error rates of semi-supervised learning methods  for CIFAR-10~\cite{krizhevsky2010convolutional} and SVHN~\cite{netzer2011reading} on 5 different folds using Wide ResNet-28-2~\cite{zagoruyko2016wide} network with different label ratios}
        \label{tab:cifar-exp}
        \vspace{6pt}
            \begin{tabular}{@{\hskip 0.05in}l@{\hskip 0.1in}l@{\hskip 0.1in}@{\hskip 0.1in}c@{\hskip 0.1in}c@{\hskip 0.1in}c@{\hskip 0.1in}c@{\hskip 0.1in}c@{\hskip 0.05in}}
            \toprule
            \multirow{2}{*}{Method}  & \multicolumn{2}{c}{CIFAR-10}& \multicolumn{2}{c}{SVHN}          \\
                         & 250 labels & 4000 labels& 250 labels & 4000 labels \\
            \midrule
            Pseudo-Label~\cite{lee2013pseudo}   &49.78$\pm$0.43 & 16.09$\pm$0.28&20.21$\pm$1.09 &7.62$\pm$0.29\\
            II Model~\cite{laine2016temporal}   &54.26$\pm$3.79 & 14.01$\pm$0.38 &18.96$\pm$1.92 &7.54$\pm$0.36 \\
            Mean Teacher~\cite{tarvainen2017mean}  &47.32$\pm$4.71 &10.36$\pm$0.25 &6.45$\pm$2.23 &3.75$\pm$0.10 \\
            VAT~\cite{miyato2018virtual}   &36.03$\pm$2.82 &13.86$\pm$0.27 &8.41$\pm$1.01 &5.63$\pm$0.20  \\
            VAT+EntMin~\cite{miyato2018virtual}   &36.32$\pm$2.13   &13.13$\pm$0.39 &8.15$\pm$0.97 &5.35$\pm$0.19 \\
            ICT~\cite{verma2019interpolation} &13.28$\pm$0.42 & 7.66$\pm$0.17  &4.31$\pm$0.17 &3.53$\pm$0.07  \\
            MixMatch~\cite{berthelot2019mixmatch}  &11.08$\pm$0.87 &6.24$\pm$0.10 &3.78$\pm$0.26 &3.27$\pm$0.31\\
            UDA~\cite{xie2019unsupervised}   &8.82$\pm$1.08 &5.29$\pm$0.25 &5.69$\pm$2.76 &2.67$\pm$0.10  \\ 
            \midrule
            UDA + Ours~(PRL)  & \textbf{7.54$\pm$0.35}  & \textbf{5.12$\pm$0.09} &\textbf{4.73$\pm$0.67}  &\textbf{2.39$\pm$0.11}  \\ 
            \bottomrule
         \end{tabular}}
      \end{table*}
         \vspace{0.5cm}
         
\subsection{Comparison with State-of-the-Art}

\paragraph{Performance on ImageNet} We first evaluate our method on the large and complex dataset ImageNet which uses 10\% of the labeled data of the dataset and treats the others as unlabeled data. We use ResNet-50~\cite{he2016deep} network as our base model to extract initial features. All the competitors in Table ~\ref{tab:imagenet} also use ResNet-50 as the backbone. It can be seen that our approach achieves new state-of-the-art performance on ImageNet~\cite{deng2009imagenet}, up to top-1 accuracy of 72.08\% and top-5 accuracy of 90.75\%. Note that UDA~\cite{xie2019unsupervised} applies the similar consistency regularization with a weak and strong augmentation techniques and obtains previous leading performance. Comparatively, we exceed UDA~\cite{xie2019unsupervised} by a margin of 3.30\% and 1.95\% in top-1 and top-5 accuracy, respectively.  Compared to the label propagation method LLP~\cite{zhuang2019local} on large-scale dataset, our approach achieves a significant improvement of 2.22\% in top-5 accuracy.

\paragraph{Performance on CIFAR-10 and SVHN} Following the standard settings, we conduct experiments on 5 different folds of labeled data and report the mean test error rate with the variance. We apply Wide-ResNet-28-2~\cite{zagoruyko2016wide} in our experiments. As shown in Table~\ref{tab:cifar-exp}, aided by the progressive GNN labeler,  we achieve  state-of-the-art performance compared with UDA~\cite{xie2019unsupervised} on CIFAR-10 and SVHN benchmarks with different label ratios. For example, with 250 labels of CIFAR-10 and SVHN, we obtain 7.54\% and 4.73\% test error rate better than UDA~\cite{xie2019unsupervised} by 1.28\% and 0.96\%, respectively. The results above significantly show the superiority and effectiveness of our PRL approach.

\section{Ablation Study}

\subsection{Effect of indegree sampler} 
In order to analyze the effect of our proposed indegree sampler, we compare with three competitors: sampler-free method (denoted as \textbf{None}) who does not select a portion of unlabeled data for training. On the contrary, it assigns a hard label to every unlabeled sample with maximum predictions; confidence-thresholding (denoted as \textbf{Conf. Thres.}) that ranks all samples by their maximum predicted probabilities and selects the samples whose maximum probability larger than a threshold of $0.5$; class-wise confidence top-k~(denoted as \textbf{Class-wise Conf. Top-k}) that ranks samples in each class independently by their probabilities of corresponding class and select k samples equally for each class~\cite{yalniz2019billion}. For fair comparison, we exclude the influence of the number of samples, and fix the number of the total selected samples by different samplers. As shown in Table ~\ref{tab:ablation_sampler}, indegree sampler achieves stably better performance than all other samplers. Especially, when the leverage our progressive GNN as labeler, indegree sampler significantly boosts the performance and outperforms the class-wise sampler by an improvement of 1.2\% referring to 72.1\% vs. 70.9\% while has a significant improvement of 2.7\% compared to the confidence thresholding sampler. It proves that our proposed sampler has more beneficial effect on searching for representative and reliable samples.

\begin{table}[h]
\centering
\caption{Comparison between different samplers on Top-1 accuracy trained by 10\%-labeled ImageNet}
\label{tab:ablation_sampler}
\setlength{\tabcolsep}{1pt}
\fontsize{9}{10}{\selectfont
\begin{tabular}{@{\hskip 0.01in}c@{\hskip 0.04in}c@{\hskip 0.04in}c@{\hskip 0.04in}c@{\hskip 0.04in}c@{\hskip 0.01in}}
\toprule
\multirow{2}{*}{Labeler} & \multicolumn{4}{c}{Sampler}          \\
        & None & Conf. Thres. & Class-wise Conf. Top-k & Ours (Indegree) \\
\midrule
CNN                     & 68.5   & 68.3          & 70.2       & 70.3 \\
\midrule
Ours  & 70.1   & 69.4          & 70.9       & \textbf{72.1} \\
\bottomrule
\end{tabular}}
\end{table}

\vspace{-10pt}
\begin{table}[h]
\center
\caption{Comparison between different labelers on Top-1 accuracy trained by 10\%-labeled ImageNet}
\label{tab:ablation_labeler}
\setlength{\tabcolsep}{1pt}
\fontsize{9}{10}{\selectfont
\begin{tabular}{@{\hskip 0.01in}c@{\hskip 0.06in}c@{\hskip  0.06in}c@{\hskip  0.06in}c@{\hskip  0.06in}c@{\hskip 0.01in}}
\toprule
\multirow{2}{*}{Sampler} & \multicolumn{4}{c}{Labeler}          \\
        & CNN & Label Propagation & GNN & Progressive GNN\\
\midrule
Ours (Indegree) & 70.3 & 70.8   &  71.2 & \textbf{72.1} \\
\bottomrule
\end{tabular}}
\end{table}

\begin{figure*}[htp]
   \centering
   \includegraphics[width=0.95\linewidth]{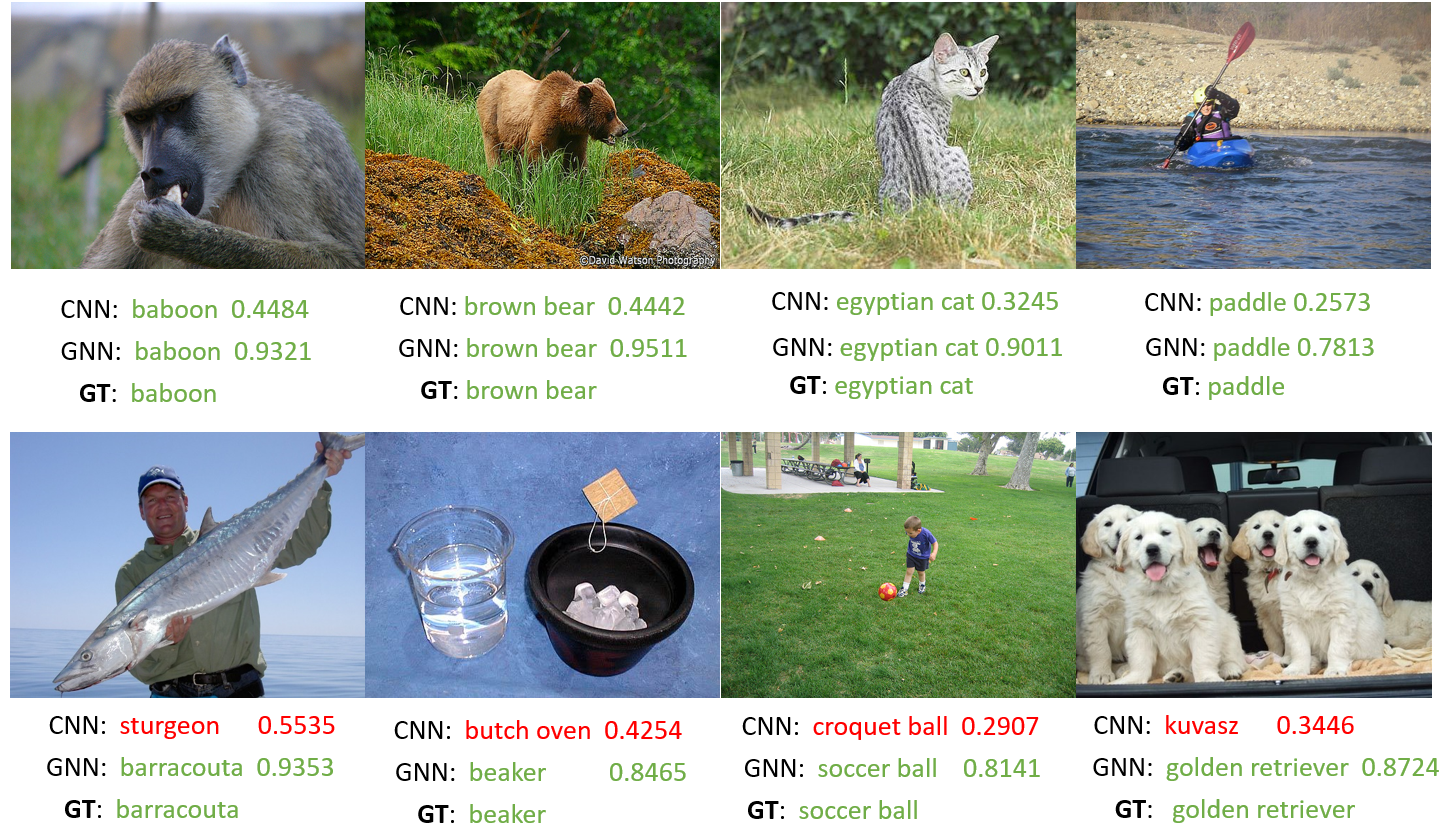}
   \caption{Visualization of examples predicted by baseline CNN model and our GNN model. We show the max probability and its corresponding class. `GT' refers to the ground-truth label. It can be shown that our GNN model helps improve the under-confident examples and correct the wrong predictions. Best viewed in color.}
   \label{fig:vis}
\end{figure*}

\begin{figure}[htb]
   \centering
   \includegraphics[width=1\linewidth]{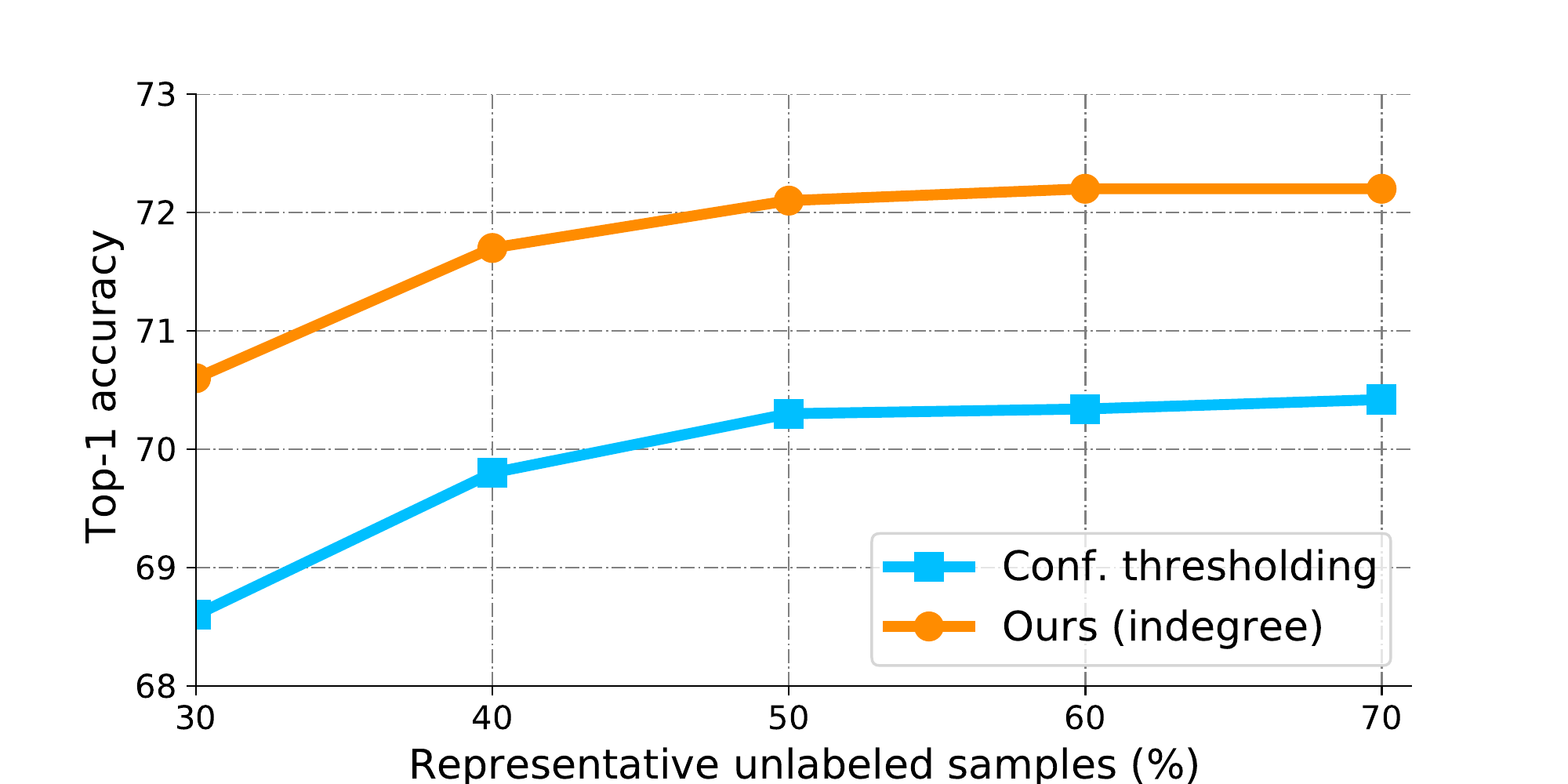}
   \caption{Progressive learning with $T=5$ iterations using different samplers}
   \vspace{1cm}
   \label{fig:sampler}
\end{figure}

\subsection{Effect of GNN labeler} 
Since we have only a small amount of labeled data for base model training, generalization ability of the model may be not reliable enough to guarantee a good performance on the unlabeled data. The pseudo-labels predicted by the base model is sometimes ambiguous or under-confident especially for the samples located near the decision boundary. Thus, the GNN labeler is proposed to incorporate the neighbor information to improve the labeling process. To verify the effectiveness of our GNN labeler, we compare two labeling methods: 1) \textbf{CNN} labeler which directly uses the one-hot labels predicted by CN N model and 2) \textbf{Label Propagation}~\cite{iscen2019label} which propagates labels from labeled data to unlabeled data via a constructed graph. We also compare between initial GNN labeler without progressive style and our completed Progressive GNN labeler. All the comparisons are based on indegree sampler who selects representative unlabeled samples for labeling.

The performance is reported in Table~\ref{tab:ablation_labeler}. Comparing to CNN labeler, GNN labeler leads to more than 0.9\% improvements on top-1 accuracy regardless of progressive operation. The possible reason is that the graph-based labeler can refine the feature and avoid introducing noisy samples. Further, label propagation can also improve the performance to 70.8\% compared to CNN labeler, which further demonstrates the power of the neighbor message. Nevertheless, GNN still possesses 0.4\% advantage over label propagation, and the dominance even becomes 1.8\% with progressive operation, indicating the effectiveness of our labeler component.

Furthermore, from Figure~\ref{fig:vis}, it can be shown that our GNN labeler boosts the performance in two aspects: Firstly, GNN can help increase the confidence for those under-confident samples; Secondly, GNN can also help correct the wrong predictions, which is presented in the second row in the figure.


\subsection{Effect of progressive learning}  
We propose a progressive learning scheme according to Section~\ref{method}. Using indegree sampler, we feed the representative samples to regularize the GNN labeler so that the labeler can provide feature with high quality to the sampler in a mutually promoting fashion. The progressive manner between labeler and sampler is repeated for $T$ iterations. To illustrate the gain of progressive learning, we conduct an ablation study on the progressive iterations. For fair comparison, the number of representative samples to expand labeled set is required to the same. At $T=0$, the labeler is trained by labeled data together with 30\% representative samples selected by indegree sampler. Every further step will constantly adds 10\% of unlabeled data size using indegree sampler and labeler. The results are presented in Figure~\ref{fig:sampler}. The accuracy rises as the progressive iteration increases and reaches the climax at $T=3$, up to 72.1\% along with 30\%, 40\% and 50\% of unlabeled sampled to expand the labeled set. Similar phenomenon can be observed when using simpler confidence thresholding sampler with threshold of $0.5$, but a lagre margin between two curves shows the superior of our GNN labeler again. Based on the climax, in our experiment, $T$ is defaultly setted to $3$. The results well verify the effectiveness of the progressive learning scheme.

\begin{table}[h]
      \center
      \setlength{\tabcolsep}{2pt}
        \caption{Analysis of the components in our learning pipeline. `finetuning' means using the consistency regularization along with the cross-entropy loss on unlabeled data and expanded labeled set. In these experiments, we use `class-wise' sampler to select pseudo-labeled data and SGC model as the labeler}
        \label{tab:steps}
        \vspace{0.5cm}
      \fontsize{9}{10}{\selectfont
         \begin{tabular}{@{\hskip 0.2in}c@{\hskip 0.25in}c@{\hskip 0.25in}c@{\hskip 0.25in}c@{\hskip 0.2in}}
            \toprule
            Labeler  & Sampler & Finetuning & Top-1 \\
            \midrule
            \xmark&   \xmark     &  \cmark      &    60.3 \\
            \xmark & \cmark   &  \cmark  & 65.1 \\
           \cmark   & \cmark   &  \xmark   & 68.5 \\
            \cmark   & \cmark  &  \cmark  & 71.3 \\
            \bottomrule
         \end{tabular}}
      \end{table}

\subsection{Benefits of learning pipeline} 
We further analyse the contributions of different components in the learning pipeline. In Table ~\ref{tab:steps}, for the CNN baseline, we directly apply finetune on the labeled data and unlabeled data. `Finetuning' means using the consistency regularization along with the cross-entropy loss on non-representative unlabeled data and expanded labeled set. As shown, we improve 4.8\% of top-1 accuracy by applying `class-wise' sampler to pseudo samples and finetuning referring to the baseline CNN. Equipped with a SGC labeler, a further gain is obtained, up to 6.2\% improvement. The result further indicates after appling the SGC labeler, more representative samples will be pseudo-labeled and added to the labeled set to help improve the generalization of model.

\section{Conclusion} 
\label{sec:conclusion}
Pseudo-labeling is a simple yet effective method for SSL. We propose a progressive representative labeling scheme to determine which part of unlabeled data to be pseudo-labeled and how to perform the pseudo-labeling. Indegree sampler is designed to select representative samples and has been verified to perform better than the confidence-thresholding sampler and the top-k confidence-ranking sampler. 
A GNN labeler cooperates with the progressive indegree sampler to refine confidence feature and improve pseudo-labeling ability. Specifically, the progressive updating GNN labeler is much more efficient than progressive updating CNN labeler. In the experiments, we demonstrate the effectiveness of the components including indegree sampler, GNN labeler, and progressive learning manner in our representative labeling.
Especially, our sampler based on kNN indegree makes essential contributions to our pipeline by selecting representative unlabeled samples. In addition, our progressive representative labeling approach is orthogonal to the consistency training approach, like UDA~\cite{xie2019unsupervised}. The Extensive experiments on deep SSL benchmarks show our state-of-the-art performance and demonstrate the superiority and the effectiveness of our method.

\newpage

\bibliography{example_paper}
\bibliographystyle{icml2021}





\end{document}